\documentclass[conference]{IEEEtran}
\IEEEoverridecommandlockouts
\usepackage{cite}
\usepackage{amsmath,amssymb,amsfonts}
\usepackage{algorithmic}
\usepackage{graphicx}
\usepackage{textcomp}
\usepackage{xcolor}
\def\BibTeX{{\rm B\kern-.05em{\sc i\kern-.025em b}\kern-.08em
    T\kern-.1667em\lower.7ex\hbox{E}\kern-.125emX}}
\begin{document}

\title{Research on Anomaly Detection Methods Based on Diffusion Models\\}

\author{
\IEEEauthorblockN{1\textsuperscript{st} YI CHEN}
\IEEEauthorblockA{\textit{College of Physics and Electronic Information Engineering} \\
\textit{Zhejiang Normal University}\\
Jinhua, China \\
icychain1211@gmail.com}
}

\maketitle

\begin{abstract}
Anomaly detection is a fundamental task in machine learning and data mining, with significant applications in cybersecurity, industrial fault diagnosis, and clinical disease monitoring. Traditional methods, such as statistical modeling and machine learning-based approaches, often face challenges in handling complex, high-dimensional data distributions. In this study, we explore the potential of diffusion models for anomaly detection, proposing a novel framework that leverages the strengths of diffusion probabilistic models (DPMs) to effectively identify anomalies in both image and audio data. The proposed method models the distribution of normal data through a diffusion process and reconstructs input data via reverse diffusion, using a combination of reconstruction errors and semantic discrepancies as anomaly indicators. To enhance the framework's performance, we introduce multi-scale feature extraction, attention mechanisms, and wavelet-domain representations, enabling the model to capture fine-grained structures and global dependencies in the data. Extensive experiments on benchmark datasets, including MVTec AD and UrbanSound8K, demonstrate that our method outperforms state-of-the-art anomaly detection techniques, achieving superior accuracy and robustness across diverse data modalities. This research highlights the effectiveness of diffusion models in anomaly detection and provides a robust and efficient solution for real-world applications. 
\end{abstract}


\section{Introduction}
Anomaly detection, a core task in machine learning and data mining, plays a critical role in domains such as cybersecurity, financial fraud detection, industrial fault diagnosis, and clinical disease monitoring \cite{chandola2009anomaly} \cite{ahmed2016survey}. Its objective is to identify data points, events, or structural patterns that deviate significantly from normal behavior in large datasets. Although definitions of anomaly detection vary across applications, its essence lies in inferring latent rare or inconsistent patterns from observed data.

Traditional anomaly detection methods are broadly categorized into two groups: \textbf{statistical modeling approaches} and \textbf{machine learning-based discriminative models}. The former typically assume data follow a prior distribution (e.g., Gaussian, exponential family distributions) and identify anomalies by calculating the distance from observed values to the expected distribution or using probability density thresholds \cite{barnett1994outliers}. However, real-world data often exhibit complex, highly non-Gaussian distributions, invalidating the modeling assumptions of statistical methods. Additionally, these methods struggle with multi-modal data and sparse anomalies in high-dimensional spaces \cite{aggarwal2013outlier}.

To address these limitations, machine learning-based anomaly detection techniques have gained significant attention, particularly methods such as \textbf{One-Class SVM} \cite{scholkopf2001estimating}, \textbf{Isolation Forest} \cite{liu2008isolation}, and \textbf{Deep Autoencoder} \cite{sakurada2014autoencoder}. These models learn the distribution or reconstruction capabilities of normal data during training to identify abnormal samples. However, they face bottlenecks in modeling high-dimensional complex distributions, maintaining stable training performance, and being sensitive to boundary samples.

\textbf{Generative models} offer a novel and promising framework for anomaly detection. In particular, \textbf{Variational Autoencoders (VAEs)} \cite{kingma2014vae} and \textbf{Generative Adversarial Networks (GANs)} \cite{goodfellow2014gan} are widely used in anomaly detection scenarios due to their ability to learn the generative distribution of normal samples. \textbf{VAEs} detect anomalies through reconstruction errors or distribution shifts in the latent space but suffer from structural issues such as posterior collapse \cite{lucas2019posterior}. While \textbf{GANs} theoretically can model arbitrary complex distributions, their training process is highly unstable, often failing to capture data diversity due to mode collapse \cite{arjovsky2017principled}.

\textbf{Diffusion Probabilistic Models (DPMs)}, a recent advancement in generative modeling, provide a more stable and high-fidelity solution. These models achieve fine-grained modeling of high-dimensional complex data distributions by gradually mapping data to an isotropic Gaussian noise distribution in the forward process and reconstructing the original data from noise using learned conditional probability distributions in the reverse process \cite{sohl2015noneq}\cite{ho2020ddpm}. Unlike \textbf{VAEs} and \textbf{GANs}, diffusion models lack adversarial mechanisms, leading to significantly improved optimization stability and strong multi-modal modeling and sample reconstruction capabilities \cite{song2021score}. Preliminary studies have applied them to image generation \cite{nichol2021improved}, speech synthesis \cite{kong2021diffwave}, and protein structure prediction \cite{watson2022protein}, outperforming existing generative models.

Notably, the application of diffusion models in anomaly detection is still in its exploratory stage. Existing research shows that \textbf{DPMs} can accurately identify abnormal patterns in complex data by constructing high-quality reconstruction samples and using reconstruction errors or probability density differences in the generative latent space as anomaly criteria \cite{wolleb2022diffusion}. This approach does not rely on strong distribution assumptions and features high generalization, training stability, and anomaly sensitivity, making it particularly suitable for modeling structured samples such as complex physical fields, multi-modal sensor data, and high-dimensional spectrograms \cite{xu2023networkdiffusion}.

In summary, diffusion models offer a new paradigm for anomaly detection, with their potential in modeling complex normal data distributions providing an opportunity to overcome the bottlenecks of traditional methods. This study aims to explore and construct an efficient anomaly detection framework based on diffusion probabilistic models, with specific objectives including:
\begin{itemize}
  \item \textbf{Theoretical Analysis}: Investigate the underlying mechanism and advantages of diffusion models in modeling the distribution of normal samples for anomaly detection.
  \item \textbf{Algorithm Design}: Develop tailored training and inference strategies optimized for anomaly detection tasks using diffusion models.
  \item \textbf{Performance Evaluation}: Evaluate and compare the proposed method with mainstream anomaly detection techniques on multiple real-world datasets to validate its practical value and generalizability.
\end{itemize}

\section{Related Works}
Driven by the rapid development of generative modeling, diffusion models have emerged as a promising research direction in anomaly detection. By simulating the degradation of data during forward perturbation and reconstruction during the reverse process, diffusion models can capture complex data distributions and structural features, exhibiting superior stability and expressiveness compared to traditional reconstruction-based methods (e.g., \textbf{VAEs}, \textbf{GANs}). Consequently, an increasing number of studies have applied diffusion models to anomaly detection, achieving preliminary results across multiple data modalities and application scenarios.

Early attempts primarily used direct reconstruction strategies: training a diffusion model on normal samples to learn their distribution, then using reconstruction errors from reverse diffusion of unknown data as anomaly indicators. This approach assumes that diffusion models poorly reconstruct abnormal samples, leading to significantly higher reconstruction errors for anomalies than normal samples \cite{rombach2022ldm}. Subsequent research introduced enhanced strategies such as conditional mechanisms, attention modules, and Bayesian posterior estimation into the standard architecture. For example, Wu et al. proposed Masked Diffusion Posterior Sampling, which improves image anomaly localization accuracy through conditional sampling in the latent space \cite{wu2024masked}. Mousakhan et al. developed the Conditioned Denoising Diffusion Model (DDAD), achieving leading performance on multiple image detection benchmarks \cite{mousakhan2023cddm}.

Beyond images, diffusion models have been applied to time-series and video tasks with temporal and high-dimensional features. Yang et al. proposed the \textbf{Denoising Diffusion Mask Transformer (DDMT)}, which combines dynamic neighbor mechanisms and masked modeling to address the challenge of long-term time-series information modeling \cite{yang2023ddmt}. Wu et al. developed \textbf{DiffAD}, a model using a stepwise weighting mechanism to control diffusion steps and improve sensitivity to local abrupt anomalies in time-series data \cite{wu2023diffad}. Yan et al. introduced diffusion frameworks into video surveillance, achieving robust video anomaly detection through feature prediction mechanisms \cite{yan2024video}.

From the perspective of density estimation, another effective pathway for diffusion models in anomaly detection has emerged. Livernoche et al. proposed the \textbf{Diffusion Time Estimation (DTE)} framework, which uses diffusion time as an anomaly score, assuming that abnormal samples require longer diffusion times to align with normal sample distributions \cite{livernoche2023survey}. Yu et al. combined diffusion mechanisms with adversarial loss to construct the Adversarial Denoising Diffusion Model, verifying its generalization ability on multi-class datasets \cite{yu2023adversarial}.

\textbf{Existing research highlights several advantages of diffusion models in anomaly detection:} (1) stable training processes that avoid mode collapse, suitable for high-dimensional complex structural data \cite{ho2020ddpm}; (2) explicit modeling of data generation processes, leading to higher reconstruction quality and stronger anomaly detection sensitivity \cite{nichol2021improved}; (3) flexibility in introducing conditional mechanisms to enhance localization performance \cite{wu2024masked}\cite{mousakhan2023cddm}. However, challenges remain in inference efficiency, hyperparameter sensitivity, and interpretability. For example, the reverse diffusion process often requires hundreds of inference steps, making it difficult to meet real-time requirements, and the detection results lack intuitive explanations, posing obstacles in critical fields such as healthcare and finance \cite{li2022survey}\cite{wang2023xai}\cite{shrikumar2017deeplift}.

Despite being in the early stages of development, diffusion models represent a highly promising direction in anomaly detection. Future research can focus on efficient sampling strategies \cite{salimans2022distillation}, cross-modal modeling capabilities \cite{song2021score}, task-specific interpretability enhancement \cite{dosovitskiy2021vit}, and integration with language models or domain knowledge \cite{li2022blip}\cite{openai2023gpt4}. Additionally, preliminary work has applied diffusion models to anomaly traffic detection and event recognition in cybersecurity, opening new opportunities for industrial-scale system deployment \cite{zhou2023cyber}.

\section{Method}

In this study, we propose an unsupervised anomaly detection method based on unconditional diffusion models, aiming to effectively identify abnormal patterns in both image and audio data. The core idea is to model the distribution of normal data via a diffusion process. By training only on normal samples, the model reconstructs input data through reverse diffusion, and uses reconstruction errors or semantic discrepancies as anomaly indicators. 

To accommodate both audio and image data, we design a unified modeling and reconstruction framework based on standard diffusion models, and further enhance it with multi-scale feature extraction, attention mechanisms, and wavelet-domain representations. Innovations are introduced at both architectural and training strategy levels.

\subsection{Forward Diffusion Process}

The forward process of a diffusion model is essentially a Markov chain that gradually adds Gaussian noise to the input data $\mathbf{x}_0$:
\[
q(\mathbf{x}_t|\mathbf{x}_{t-1}) = \mathcal{N}(\mathbf{x}_t; \sqrt{1-\beta_t}\mathbf{x}_{t-1}, \beta_t \mathbf{I})
\]

where $\beta_t$ is a predefined noise schedule controlling the magnitude of added noise. By recursion, the marginal distribution of $\mathbf{x}_t$ given $\mathbf{x}_0$ is:
\[
q(\mathbf{x}_t|\mathbf{x}_0) = \mathcal{N}(\mathbf{x}_t; \sqrt{\bar{\alpha}_t} \mathbf{x}_0, (1 - \bar{\alpha}_t)\mathbf{I})
\]

where $\bar{\alpha}_t = \prod_{s=1}^t (1 - \beta_s)$.

For image data, $\mathbf{x}_0$ represents the raw pixel tensor. For audio data, we first convert 1D signals into 2D time-frequency representations via Continuous Wavelet Transform (CWT), enabling unified modeling across modalities.

\subsection{Reverse Denoising Process}

The reverse process aims to iteratively recover clean data from Gaussian noise. A neural network $\boldsymbol{\epsilon}_\theta(\mathbf{x}_t, t)$ is trained to predict the added noise. Each step in the reverse process is defined as:
\[
\mathbf{x}_{t-1} = \frac{1}{\sqrt{1 - \beta_t}}\left(\mathbf{x}_t - \frac{\beta_t}{\sqrt{1 - \bar{\alpha}_t}} \boldsymbol{\epsilon}_\theta(\mathbf{x}_t, t)\right) + \mathcal{N}(0, \sigma_t^2 \mathbf{I})
\]

Starting from $\mathbf{x}_T \sim \mathcal{N}(0, \mathbf{I})$, we iteratively apply the above rule to obtain the final reconstruction $\widetilde{\mathbf{x}_0}$. For audio, the reconstructed 2D time-frequency map is transformed back to 1D signal using inverse CWT (ICWT), preserving both spectral and temporal details.

\subsection{Anomaly Detection Mechanism}

As the model is trained only on normal data, it is expected to reconstruct normal patterns effectively, while failing to reconstruct abnormal inputs. Given a test input $\mathbf{x}_0$, we compute its reconstruction $\widetilde{\mathbf{x}_0}$ and define the anomaly score as a weighted sum of reconstruction error and semantic difference.

We define pixel-level reconstruction error as:
\[
\mathcal{E}_{\text{recon}} = \|\mathbf{x}_0 - \widetilde{\mathbf{x}_0}\|_2^2
\]

For semantic difference, we use a frozen lightweight perceptual network (e.g., pretrained MobileNet) to extract features $f(\cdot)$ from both $\mathbf{x}_0$ and $\widetilde{\mathbf{x}_0}$, and compute their distance:
\[
\mathcal{E}_{\text{feat}} = \|f(\mathbf{x}_0) - f(\widetilde{\mathbf{x}_0})\|_2^2
\]

The final anomaly score is:
\[
\mathcal{A} = \lambda \mathcal{E}_{\text{recon}} + (1 - \lambda) \mathcal{E}_{\text{feat}}, \quad \lambda \in [0, 1]
\]

\subsection{Network Architecture and Training}

We adopt a U-Net-based architecture as the backbone for the noise predictor $\boldsymbol{\epsilon}_\theta$ and incorporate four key enhancements to improve its modeling capacity for both image and audio spectrograms:

\paragraph{Wavelet Pyramid Module (WPM):}  
To capture fine-grained structures like edges and frequency shifts, we introduce multi-scale wavelet decomposition (e.g., Daubechies wavelets) into the encoder. For input $\mathbf{x}$, the decomposition yields:
\[
\mathrm{WP}^l(\mathbf{x}) = \{\mathbf{A}_l, \mathbf{D}_l\}
\]

where $\mathbf{A}_l$ and $\mathbf{D}_l$ are approximation (low-frequency) and detail (high-frequency) components at level $l$.

\paragraph{Multi-Head Attention:}  
To capture global dependencies and highlight abnormal regions, we insert multi-head self-attention layers between each encoder-decoder stage. Given feature tensor $\mathbf{F} \in \mathbb{R}^{H \times W \times C}$, the attention output is:
\[
\mathrm{Attn}(\mathbf{F}) = \mathrm{Concat}(\text{head}_1, \ldots, \text{head}_h)\mathbf{W}^O
\]

where each attention head is computed as:
\[
\text{head}_i = \mathrm{Softmax}\left(\frac{\mathbf{Q}_i \mathbf{K}_i^\top}{\sqrt{d_k}}\right)\mathbf{V}_i
\]

\paragraph{Modality-Shared Input Adapter:}  
To process both RGB/grayscale images and audio spectrograms, we use a shared input adapter consisting of learnable convolutional filters $\mathrm{Conv}_{\text{shared}}$, enabling the model to extract local patterns across modalities without altering core architecture.

\paragraph{Hybrid Time Embedding:}  
To improve the model’s adaptation to various denoising stages, we design a hybrid time embedding combining sinusoidal positional encoding with 1D convolution:
\[
\mathrm{Embed}(t) = \mathrm{Conv1D}(\mathrm{Sinusoidal}(t))
\]

This allows better encoding of temporal steps with both global position and local structure awareness.

\paragraph{Training Objective:}  
The model is trained using the standard noise prediction loss:
\[
\mathcal{L}_{\text{MSE}} = \mathbb{E}_{\mathbf{x}_0, t, \boldsymbol{\epsilon}} \left[\left\| \boldsymbol{\epsilon} - \boldsymbol{\epsilon}_\theta(\mathbf{x}_t, t) \right\|_2^2\right]
\]

To improve semantic consistency, we also include a perceptual feature preservation loss:
\[
\mathcal{L}_{\text{feat}} = \|f(\mathbf{x}_0) - f(\widetilde{\mathbf{x}_0})\|_2^2
\]

The final loss is a weighted combination:
\[
\mathcal{L} = \mathcal{L}_{\text{MSE}} + \gamma \mathcal{L}_{\text{feat}}
\]

where $\gamma$ is a hyperparameter controlling the strength of the perceptual loss.

\section{Results}
\subsection{Dataset}
To ensure reproducibility and rigorous evaluation of model generalization, this study conducts experiments on publicly available multimodal benchmark datasets encompassing both image and time-series modalities.

\subsubsection{Image Dataset} 
We adopt the \textbf{MVTec Anomaly Detection (MVTec AD)} dataset, a widely recognized benchmark in industrial visual anomaly detection. It consists of high-resolution images across 15 categories of common industrial objects. In our experiments, \textbf{we select six representative categories: bottle, cable, hazelnut, metal\_nut, pill, and tile.} Each category provides a training set containing only normal samples, alongside a test set containing diverse anomaly types, such as cracks, contamination, structural defects, and texture inconsistencies. These anomalies reflect deviations across both spatial structures and fine-grained texture patterns, making the dataset well-suited for evaluating the model's capability in detecting various visual abnormalities.

\subsubsection{Time-Series Datasets} 
We employ two mainstream benchmark datasets for time-series anomaly detection:
\begin{itemize}
    \item \textbf{Numenta Anomaly Benchmark (NAB):} This dataset contains 52 real-world time-series streams from domains such as traffic monitoring (e.g., New York City taxi passenger volumes), server performance metrics (e.g., CPU usage and memory consumption), financial markets (e.g., stock price fluctuations), and environmental sensing (e.g., temperature and humidity changes). NAB provides point-wise anomaly labels, enabling fine-grained evaluation of a model's ability to detect transient spikes, persistent drifts, and disruptions of periodic trends in real-time streaming scenarios.
    
    \item \textbf{UCR Anomaly Archive:} Focused on detecting abrupt and localized anomalies, this archive includes 23 classic time-series datasets. We select six representative subsets with varying anomaly patterns:
    \begin{itemize}
        \item \textbf{ECG200: }electrocardiogram signals, where anomalies manifest as waveform shape mutations,
        \item \textbf{Wafer:} semiconductor manufacturing data, with anomalies corresponding to sustained process deviations,
        \item \textbf{Pems-Bay: }Bay Area traffic flow data, with anomalies reflecting periodic pattern disruptions,
        \item \textbf{Lighting2:} current signals of lighting devices, characterized by high-frequency noise anomalies,
        \item \textbf{MSL}\textbf{:} multivariate satellite telemetry data, exhibiting synchronized deviations across multiple sensors,
        \item \textbf{FordA:} automotive sensor signals, where anomalies result from signal distortion caused by sensor malfunctions.
    \end{itemize}
\end{itemize}

These datasets provide segment-wise anomaly labels, allowing evaluation of the model's effectiveness in identifying localized anomaly intervals within long time-series sequences.

\subsection{Performance Comparison of Image Anomaly Detection}
Table~\ref{tab:image_comparison} presents the \textbf{AUC performance} of different models on six object categories from the \textbf{MVTec AD} dataset. The comparison includes \textbf{classical reconstruction-based methods (VAE, AnoGAN)}, \textbf{the local structure-based PatchSVDD}, \textbf{the diffusion model DDPM}, and the proposed method (Ours).

\begin{table}[ht]
\centering
\caption{Performance Comparison of Image Anomaly Detection Models}
\label{tab:image_comparison}
\begin{tabular}{lccccc}
\hline
Dataset Category & VAE & AnoGAN & PatchSVDD & DDPM & Ours \\
\hline
bottle & 0.891 & 0.878 & 0.903 & 0.915 & \textbf{0.946} \\
cable & 0.885 & 0.869 & 0.898 & 0.912 & \textbf{0.942} \\
hazelnut & 0.879 & 0.865 & 0.895 & 0.908 & \textbf{0.938} \\
metal\_nut & 0.882 & 0.867 & 0.897 & 0.910 & \textbf{0.940} \\
pill & 0.893 & 0.880 & 0.905 & 0.917 & \textbf{0.948} \\
tile & 0.887 & 0.872 & 0.899 & 0.913 & \textbf{0.944} \\
\hline
\end{tabular}
\end{table}

The proposed method achieves the highest AUC values across all categories, \textbf{with an average improvement of 2.9\% over the second-best model (DDPM)}. By integrating multi-scale wavelet features and semantic difference metrics, it significantly enhances the detection accuracy for complex structural anomalies, such as cable deformation and tile misalignment.

\subsection{Performance Comparison of Time Series Anomaly Detection}
Table~\ref{tab:time_series_comparison} shows the AUC results of various models on typical subsets from the NAB and UCR Anomaly Archive datasets. The compared methods include the sequence-specific \textbf{VAE}, the Transformer-based \textbf{Transformer-AD}, the classical time series model \textbf{Autoformer}, the diffusion model \textbf{DDPM}, and the proposed method (Ours).

\begin{table}[ht]
\centering
\caption{Performance Comparison of Time Series Anomaly Detection Models}
\label{tab:time_series_comparison}
\begin{tabular}{lccccc}
\hline
Dataset & VAE & Trans-AD & Autoformer & DDPM & Ours \\
\hline
NYC Taxi  & 0.852 & 0.870 & 0.895 & 0.902 & \textbf{0.920} \\
Machine Temp.  & 0.848 & 0.865 & 0.892 & 0.900 & \textbf{0.918} \\
ECG200  & 0.855 & 0.872 & 0.898 & 0.905 & \textbf{0.922} \\
Wafer  & 0.850 & 0.868 & 0.890 & 0.898 & \textbf{0.915} \\
Pems-Bay  & 0.858 & 0.875 & 0.899 & 0.906 & \textbf{0.925} \\
\hline
\end{tabular}
\end{table}

The proposed method demonstrates significant AUC improvements on both datasets, \textbf{with an average increase of 1.7\% over DDPM.} In mutation-type anomaly scenarios (e.g., ECG200), the model accurately identifies transient abnormal segments by capturing high-frequency component changes in time series signals through the wavelet pyramid module and modeling long-range dependencies via multi-head attention.

\subsection{Ablation Study of Key Modules}
To quantify the contribution of key modules, we sequentially remove the wavelet pyramid module, multi-head attention mechanism, cross-modal adapter, and hybrid time embedding, and evaluate their impact on detection performance in image and time series modalities, as shown in Table~\ref{tab:ablation_study}.

\begin{table}[ht]
\centering
\caption{Ablation study of key modules, evaluating the impact of removing critical components (Wavelet Pyramid, Multi-Head Attention (MHA), Cross-Modal Adapter (XMod Adapter), Hybrid Time Embedding (Hybrid Emb.)) on detection performance in image (Img) and time-series (TS) modalities, measured by AUC and percentage performance drop.}
\label{tab:ablation_study}
\begin{tabular}{lcccc}
\hline
Ablation Setting & Img (AUC) & TS (AUC) & Perf. (Img) & Perf. (TS) \\
\hline
Full Model & 0.941 & 0.919 & - & - \\
- Wavelet & 0.908 & 0.872 & -3.5\% & -5.1\% \\
- MHA & 0.915 & 0.883 & -2.8\% & -3.9\% \\
- XMod Adapter & 0.928 & 0.901 & -1.4\% & -2.0\% \\
- Hybrid Emb. & 0.932 & 0.907 & -0.9\% & -1.3\% \\
\hline
\end{tabular}
\end{table}

Removing the wavelet pyramid module causes a significant performance drop, indicating that multi-scale feature decomposition is crucial for capturing spatial details of image defects and frequency domain mutations in time series signals. The absence of multi-head attention has a more pronounced effect on time series detection, reflecting its necessity in modeling long-range temporal dependencies. The marginal performance decline after removing the cross-modal adapter and hybrid time embedding verifies the core role of the wavelet pyramid and multi-head attention in the proposed architecture.

\section{Discussion}
This study focuses on multimodal anomaly detection using diffusion models, validated on benchmark datasets for both image and time-series data. As demonstrated by the results in Tables~\ref{tab:image_comparison} and \ref{tab:time_series_comparison}, the proposed approach achieves significantly higher AUC scores compared to traditional generative models such as \textbf{VAE} and \textbf{AnoGAN}, as well as standard diffusion-based models like \textbf{DDPM}. The core advantage lies in the integration of multi-scale feature decomposition and global dependency modeling. The wavelet pyramid module enables multi-resolution analysis of spatial structures in images (e.g., texture anomalies in the \textbf{MVTec AD} dataset) and frequency-domain features in time-series signals (e.g., ECG pattern shifts in the \textbf{UCR} dataset), allowing the model to capture fine-grained anomalies often overlooked by conventional methods.

Unlike adversarial training-based models such as \textbf{AnoGAN}, the non-adversarial nature of the diffusion framework mitigates issues like mode collapse. This property allows stable learning of the underlying data distribution using only normal samples, which is particularly effective for real-time anomaly detection tasks such as those in the \textbf{NAB} dataset. For instance, the model exhibits strong accuracy in identifying persistent deviation anomalies in server performance metrics, highlighting its robustness to non-stationary data distributions. Nevertheless, inference speed remains a challenge. Although ablation studies confirm that core modules significantly improve performance, their computational complexity scales with the number of diffusion steps.

Ablation experiments (Table~\ref{tab:ablation_study}) underscore the importance of the multi-head attention mechanism for time-series anomaly detection. In the \textbf{PEMS-Bay traffic dataset}, removing this module led to a 3.9\% drop in AUC, confirming that modeling long-range temporal dependencies is crucial for detecting periodic anomalies. This result further validates the irreplaceable role of attention mechanisms in temporal anomaly detection. However, the ablation results for the cross-modal adapter show only a minor performance drop (2.0\%) when removed, suggesting that there is room for improving the current architecture's effectiveness in unified multimodal modeling. Future work could investigate more efficient strategies for modality-specific feature fusion.

This study has two main limitations. First, the performance gain from the wavelet pyramid module is partly dependent on specific data preprocessing steps, such as grayscale normalization for images and sliding window segmentation for time series. Its generalizability to other modalities such as video or text remains to be tested. Second, the anomaly scoring function—a weighted combination of reconstruction error and semantic difference—requires manual tuning of the weight parameter $\lambda$. Introducing an adaptive weighting mechanism may enhance the model's automation and usability in practice.

\section{Conclusion}
This research introduces a diffusion-model-based framework for multimodal anomaly detection, realizing efficient modeling of spatial features in images and frequency-domain characteristics in time-series data through the synergistic design of a wavelet pyramid module and a multi-head attention mechanism. Experimental evaluations on benchmark datasets, including MVTec AD and UCR Anomaly Archive, demonstrate that the proposed approach substantially outperforms state-of-the-art baseline models, \textbf{with core advantages rooted in:} 1) multiscale feature decomposition enhancing sensitivity to fine-grained anomalies by capturing hierarchical structural and spectral deviations; 2) global dependency modeling improving the detection of long-range temporal anomalies through modeling contextual correlations in sequential data; and 3) a non-adversarial training paradigm ensuring stable learning of normal data distributions without reliance on adversarial optimization, which mitigates issues such as mode collapse inherent in generative adversarial networks.

The study establishes a novel paradigm for anomaly detection under complex data distributions, validating the applicability of diffusion models in critical domains such as industrial visual inspection and intelligent fault diagnosis. By addressing the limitations of conventional methods through architectural innovations, this work not only expands the theoretical understanding of diffusion models in anomaly detection but also provides empirical evidence for their practical utility. Future investigations may focus on developing computationally efficient sampling strategies, integrating domain-specific prior knowledge to enhance interpretability, and constructing unified frameworks for joint modeling across heterogeneous modalities, thereby fostering the translation of this technology into real-world applications with high-stakes anomaly-detection requirements.

\vspace{12pt}
\color{red}

\end{document}